%% file: corl.tex
\documentclass{article}

\usepackage[preprint]{template/corl_2026}
\usepackage{graphicx}
\usepackage{booktabs}
\usepackage{amsmath,amssymb}
\usepackage{enumitem}

\newcommand{\best}[1]{\textbf{#1}}
\newcommand{\na}{--}
\newcommand{\ckpt}[1]{} 

\title{From Teleoperation to Long-Horizon Dexterity: Evaluating Visuomotor Skills and Offline Proxies for Dexterous Manipulation}

\author{
  Ammar N. Abbas$^{*}$, Aayush Jain$^{*}$, Prashanth Viswanath,\\
  Nivesh Gadipudi, Giulia Vilone, Jim Gibbons\\
  Analog Devices, Inc., EdgeAI Robotics, Limerick, Ireland\\
  $^{*}$Equal contribution. Contact: \texttt{ammar.abbas@analog.com}
}

\begin{document}
\maketitle

\begin{figure}[h!]
\centering
\includegraphics[width=\textwidth]{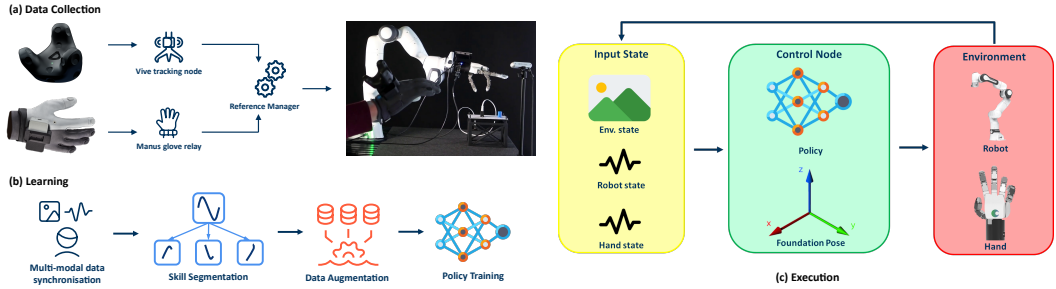}
\caption{Skill-orchestration workflow: (a) teleoperated data collection,
(b) segmentation into motion primitives and learned skills, (c) execution, where
the learned skill acts after classical point-to-point motion.}
\label{fig:system}
\end{figure}

\begin{abstract}
Imitation learning policies are intended to make robotic manipulation robust to variations in object pose, appearance, and environment that challenge pre-programmed motion. However, real-world performance remains difficult to predict, as conventional optimization signals such as validation loss show no consistent relationship with robot success. To identify the design choices that govern real-world performance, 281 imitation-learning policies spanning seven policy families were evaluated in approximately 1,700 real-robot trials of RJ45 connector grasping and insertion using a five-fingered robotic hand. Controlled ablations showed that observation and action representations dominated conventional optimizer and sampling choices. For point-cloud policies, region-of-interest cropping, representation in the robot-base frame, and normalization within the point encoder each improved success by 70–100 percentage points. For the vision-language-action policy, absolute rather than relative action prediction improved insertion success by up to 70 percentage points. Skill decomposition, multi-view perception, and recovery demonstrations were associated with the strongest-performing systems, although their effects were not independently isolated. Of 67 offline metrics evaluated, only five measures of predicted-motion smoothness and directness reliably correlated with insertion success (\(r=-0.70\) to \(-0.80\)), and only within a single policy family. These results identify representation choices as key determinants of real-world imitation-learning performance and show that offline metrics can shortlist policies, but cannot replace real-robot evaluation.\end{abstract}


\section{Introduction}
Imitation learning offers a practical route to robotic manipulation that
generalises across the variations in object pose, appearance and environment
that defeat pre-programmed motion, learning a direct observation-to-action
mapping from demonstrations rather than hand-engineered
controllers~\cite{argall2009survey,zhao2023act,chi2024diffusion}. Recent
visuomotor policies including diffusion and flow-based action generators, transformer
decoders, and large vision-language-action models, have driven rapid progress
on real single-arm tasks~\cite{chi2024diffusion,zhao2023act,black2024pi0}. Yet
the paradigm inherits well-known liabilities: behavioral cloning accumulates
error under co-variate shift~\cite{ross2011dagger}, and real-world success
remains hard to predict because conventional offline signals such as validation
loss correlate weakly, or not at all, with task
performance~\cite{mandlekar2021matters}. As a result, the design choices that
govern a policy, like observation and action representations, encoder pretraining,
sampling and optimization, are typically settled by expensive real-robot
trials, and systematic evidence to guide them is scarce, especially for the
dexterous, contact-rich skills demanded by multi-fingered
hands~\cite{florence2021implicit,openai2020dexterous}.

We study these questions on a representative contact-rich task: inserting an
RJ45 cable connector, which demands millimetre precision on a deformable object
where small changes in cable pose or grasp break fixed trajectories and still
challenge end-to-end learned
policies~\cite{chi2024diffusion,ze2024dp3,black2024pi0}. We decompose the task
into four stages: two moves between known poses handled by classical motion,
and two learned contact-rich skills, \emph{grasp} and \emph{insert}
(Section~\ref{sec:system}). We ask how best to build those skills for a
five-fingered hand. We contribute (i)~a real-robot ablation study that isolates
individual policy components (data augmentation, image-encoder pretraining,
action representation, point-cloud reference frame (camera or robot base),
cropping and normalization, prediction horizon, optimizer, denoising sampler and
policy family) and reports for every comparison all training settings that
changed between the two runs; and (ii)~a leak-free test of whether offline
metrics can stand in for robot trials.

\section{System and Protocol}
\label{sec:system}

\textbf{Preliminaries.} Every policy we compare maps the current observation and
robot state to a short chunk of future actions, differing along three axes.
\emph{Observation encoder:} Diffusion Policy (DP)~\cite{chi2024diffusion} and
ACT~\cite{zhao2023act} encode RGB images; DP3~\cite{ze2024dp3} encodes a point
cloud with a small PointNet; and iDP3~\cite{ze2024idp3} applies a multi-stage
point encoder to egocentric clouds. \emph{Action generator:} DP, DP3 and iDP3
denoise the action chunk, flow matching~\cite{lipman2023flow} instead regresses
a velocity field, ACT decodes the chunk with a conditional-autoencoder
transformer, and an equivariant diffusion policy~\cite{yang2024equibot} bakes
rotation and scale symmetry into its point encoder. \emph{Pretraining:} these
policies are trained on our demonstrations, at most with an ImageNet- or
R3M-initialised image encoder, whereas the vision-language-action (VLA) model
$\pi_{0.5}$~\cite{pi05,black2024pi0} is pretrained on large cross-embodiment
data and finetuned with LoRA~\cite{hu2022lora}. We selected these policies as widely used, open-source methods with strong
reported results on real single-arm manipulation from RGB or depth input.
Comparable ablations are typically confined to simulation or parallel-jaw
grippers, and cable manipulation has largely relied on engineered
pipelines~\cite{zhu2023cable}.

\textbf{Setup.} An RJ45 cable connector sits in a socket on one of two grasp
jigs, at the centre or left of the work area, while a third jig holds the
target socket. Each trial proceeds in four stages (Appendix~\ref{app:setup}):
(1)~classical motion to a preset start pose at the jig holding the cable, which
FoundationPose~\cite{wen2024foundationpose} identifies; (2)~a learned policy
grasps the connector and pulls it from its socket; (3)~classical motion carries
it to the insert jig; and (4)~a learned policy inserts it. Demonstrations are
teleoperated on a Franka Research~3 arm with a Tesollo DG5F-M hand at 30\,Hz;
policies run at 10\,Hz, since 10\,Hz replay reproduced the outcomes of 30\,Hz
replay. Policies observe
a wrist RGB-D camera, and \emph{two-view} variants add a static third-person
camera; codebases and settings are in Appendix~\ref{app:policies}.

\textbf{Scoring.} Each trial is scored by the last stage the robot completed:
Fail ($F$, the hand did not close on the connector), Grasp ($G$, it closed but
never reached the grip threshold needed to lift), Pick ($P$, pulled out but not
seated) or Insert ($I$, seated in the target socket). An inserted trial was
therefore also picked, so over $n$ trials Pick${}=(I{+}P)/n$,
Insert$\,|\,$pick${}=I/(I{+}P)$ and Insert${}=I/n$, pooled per checkpoint with 95\% Clopper--Pearson intervals; the rates of two
policies that differ by one change are compared with Fisher's exact test.

\section{Ablation Results}

\begin{figure}[t]
\centering
\begin{minipage}[b]{0.57\textwidth}
\centering
\includegraphics[width=\linewidth]{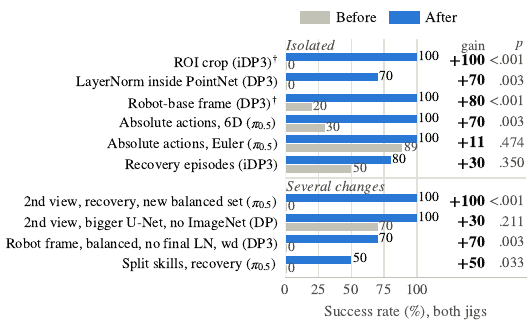}\\[-1pt]
{\small (a)}
\end{minipage}\hfill
\begin{minipage}[b]{0.4\textwidth}
\centering
\includegraphics[width=\linewidth]{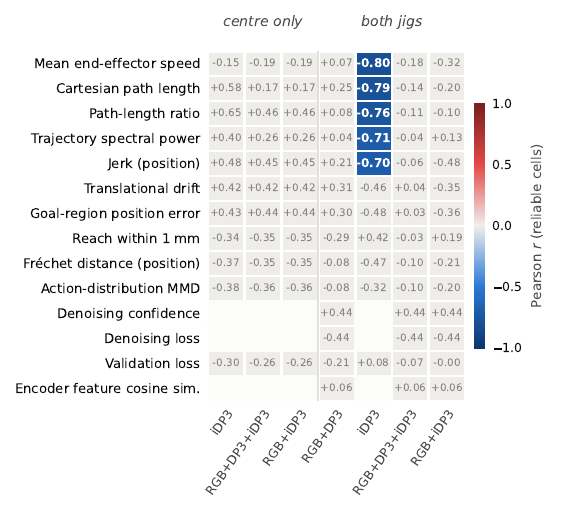}\\[-1pt]
{\small (b)}
\end{minipage}
\caption{\textbf{(a)} Before/after success of paired runs, both jigs, with
Fisher exact $p$; labels name every setting that differed. $^{\dagger}$Pick rate
of the grasp skill; other rows: insert rate of the grasp-then-insert pairing.
\textbf{(b)} Pearson $r$ between offline metrics and the insert-given-pick rate,
one metric per cell within one policy group (centre-only or both-jig
checkpoints); coloured cells are reliable.}
\label{fig:results}
\end{figure}

\textbf{Paired comparisons.} Fig.~\ref{fig:results}a compares pairs of policies that differ in one component
(\emph{isolated}) or, where marked, several; Appendix~\ref{app:changes} lists
every change tested, and rates are percentages of all pooled trials
(Appendix~\ref{app:protocol}). Among isolated changes, a region-of-interest
(ROI) crop raised the iDP3 pick rate from 0 to 100\%, and applying PointNet's
layer normalisation after every hidden layer, rather than only to its output,
raised DP3 insertion from 0 to 70\%. Predicting absolute instead of relative
actions raised the two-view $\pi_{0.5}$ insert from 30 to 100\% with the 6D
rotation representation~\cite{zhou2019continuity}, but only from 89 to 100\%
with Euler angles. On the same balanced dataset (Appendix~\ref{app:data}), the robot-base frame raised PointNet picks from 20 to 100\%, with a second pair on
other clouds showing the same trend. Recovery episodes alone raised the iDP3
pipeline from 50 to 80\%, which is not significant at this $n$.

The largest gains came from several changes at once. A second camera view
together with recovery episodes (Appendix~\ref{app:data}), on a balanced set
re-drawn from the two-camera recordings (180 episodes plus 108 recovery
episodes, against 186), raised the $\pi_{0.5}$ insert from 0 to 100\%, while a second view with a larger UNet and no ImageNet
initialisation raised the RGB DP insert from 70 to 100\% on the same data. Every
grasp-and-insert policy inserted in at most 10\% of trials, whereas the best decomposed pipelines inserted in every trial, and the $\pi_{0.5}$ decomposed
pipeline with recovery episodes rose from 0 to 50\%. The split is therefore
consistently associated with insertion, though its effect is never isolated from
other changes.

\textbf{Second-order factors.} No visual augmentation improved the RGB grasp
skill's pick rate over the unaugmented baseline (40\%). ImageNet initialisation helped grasping, most notably at the left jig, but not
insertion, where pretrained and from-scratch backbones each inserted 50--70\%. And
none of
fifteen perturbations of the iDP3 grasp skill (optimiser, width, depth,
DPM-Solver++~\cite{lu2022dpmpp}, $v$-prediction~\cite{salimans2022progressive})
clearly beat the defaults.

\textbf{Families, cost and robustness.} In Table~\ref{tab:family}, iDP3,
$\pi_{0.5}$ and a two-view DP insert pipeline achieve insertion in every trial,
RGB DP and PointNet DP3 in 70\%, and the best flow-matching and ACT pipelines in
30--50\%, while most PointNet DP3 and ACT pairings never inserted. The compact
point-cloud skill matches the VLA at an eighth of its inference memory. The iDP3 pipeline retained 80--100\% of its
insertions under changed lighting, background, site and sunlight; only a
tighter-tolerance socket broke it (0--40\%; Table~\ref{tab:app-robust}).

\begin{table}[t]
\centering
\caption{Best configuration per family, both jigs: pick and insert-given-pick
rates [95\% CI], best pairing's insert rate, median over the family's pairings
(Typical, count), parameters of the larger skill ($\pi_{0.5}$: total,
LoRA-finetuned), summed grasp+insert training time per epoch on four H100s, and
larger per-GPU memory at batch 32 (training) / batch 1 (inference), which at
equal batch scales mainly with model size. $^{\ddagger}$Per step (30k per
skill). $^{\dagger}$Insert only, after the $\pi_{0.5}$ grasp. $^{p}$three
picks.}
\label{tab:family}
\footnotesize
\setlength{\tabcolsep}{2.0pt}
\begin{tabular*}{\linewidth}{@{\extracolsep{\fill}}lccccccc@{}}
\toprule
Family (best config.; obs.) & Pick [CI] & Ins.$|$pick [CI] & Insert & Typical & Params & s/ep & GB tr./inf. \\
\midrule
DP, ImageNet, 6D; RGB\ckpt{07.21 / 07.25}                   & \best{100} [95,100] & 82 [48,98]  & 70 & 60 (5)  & 72\,M & 71 & 18/1.1 \\
DP two-view insert$^{\dagger}$; 2 RGB\ckpt{06.22 grasp / 07.23-05.23}  & 98 [94,100] & \best{100} [69,100] & \best{100} & \na & 281\,M & 208 & 34/2.1 \\
DP3 PointNet, robot frame; PCD\ckpt{09.11-21.37 / 09.11-18.37} & 90 [55,100] & 78 [40,97]  & 70 & 0 (26)  & 67\,M & 18 & 5.2/1.0 \\
iDP3, ROI crop; PCD\ckpt{08.13 / 08.16}            & \best{100} [95,100] & \best{100} [69,100] & \best{100} & 57 (22) & 69\,M & 18 & 6.0/1.0 \\
$\pi_{0.5}$ LoRA, 2-view insert; 2 RGB\ckpt{06.22 / 07.06}  & 98 [94,100] & \best{100} [69,100] & \best{100} & 10 (20) & 3.3\,B & 1.6$^{\ddagger}$ & 24/8 \\
FM ResNet; RGB\ckpt{08.26-23.05 pair}                          & \best{100} [69,100] & 40 [12,74] & 40 & 40 (1)  & 260\,M & 66 & 21/1.9 \\
FM PointNet, $H{=}16$; PCD\ckpt{08.29 pair}                    & 80 [44,97]  & 62 [24,91]  & 50 & 15 (4)  & 255\,M & 26 & 10/1.9 \\
FM iDP3 enc., $H{=}16$; PCD\ckpt{08.23-00.59 / 08.25-16.09 both-jig pair}        & 73 [39,94] & 50 [12,88] & 33 & 0 (3) & 69\,M & 26 & 5.5/1.0 \\
ACT ResNet (best of 13); RGB\ckpt{09.03-11.42 / 13.49; times from lt_train.log, 1 GPU}         & 30 [7,65]   & 100$^{p}$ [29,100] & 30 & 0 (3) & 84\,M & 80 & 21/1.2 \\
\bottomrule
\end{tabular*}
\end{table}
\section{Offline Metrics as Proxies for Robot Trials}

For every checkpoint with robot trials we computed 67 metrics on validation
episodes held out from its own training set (list, selection criteria and loss
curves in Appendix~\ref{app:offline}, Fig.~\ref{fig:app-loss}) and asked whether any tracked the checkpoint's robot success rate. Each \emph{cell} of Fig.~\ref{fig:results}b correlates one metric within one
group of policy families and one jig set (centre only or both).
A cell is reliable only if its correlation survives correction for multiple
testing and keeps its sign when any single policy is removed. No cell passed for
the centre-only checkpoints or for groups mixing RGB and point-cloud policies,
whose correlations were weak or sign-flipped depending on the policies
included, so these metrics cannot compare policy types.

Only for the 21 iDP3 insert checkpoints tested at both jigs did five of the 67
pass, all negative ($r{=}{-}0.70$ to ${-}0.80$): lower values inserted more
often. The five describe the predicted end-effector path over 50 frames per
validation episode (mean speed, path length, path-length ratio, jerk and
spectral power), so all are low when the path is short, direct and smooth. A policy whose
predictions jitter frame to frame traces a long, jerky path that appears on the
robot as hovering and misalignment at the socket. Because the five correlate
strongly with each other ($r{=}0.82$--$0.99$), they carry the same information
and act as a single smoothness score that can rank a family's
insert checkpoints before robot testing. No metric tracked the pick rate, plausibly because grasp failures arise at first
contact, which replayed observations do not reproduce.

\section{Conclusion and Takeaways}
Across ~1,700 real-robot trials and 281 policies, a compact point-cloud
policy, iDP3 with a region-of-interest crop, inserted the RJ45 connector in every trial on both jigs, matching a finetuned VLA at an eighth of its inference memory, and held 80--100\% of insertions under environment variation. Four takeaways:

\begin{itemize}[leftmargin=*, nosep]
  \item \textbf{Representation over tuning.} Observation and action choices: ROI
  cropping, robot-base frames, in-encoder normalisation, absolute actions with 6D
  rotations, shifted success by 70--100 points, far more than the optimiser or
  sampler.
  \item \textbf{Decompose contact-rich tasks.} Separate learned \emph{grasp} and
  \emph{insert} skills bracketed by classical motion beat end-to-end policies,
  which inserted in at most 10\% of trials; each skill also wants its own data:
  the grasp skill, which must cope with varying start positions, did best on 
  balanced and diverse data, the precision-bound insert skill on all
  episodes.
  \item \textbf{Offline metrics shortlist, not replace.} An offline smoothness
  score ranks insert checkpoints within one family, but no metric generalised
  across policy types or tracked the pick rate.
  \item \textbf{Compact can match large.} A small point-cloud skill rivals a
  pretrained VLA here, a favourable trade under memory and compute limits.
\end{itemize}

\acknowledgments{We thank the extended EdgeAI robot team in Limerick and the
collaborating software team in Philippines, and the NVIDIA, Tesollo, Franka Robotics, and MANUS teams for their
support.}

\bibliography{references}

\clearpage
\appendix
\input{appendices/appendix_corl}

\end{document}

%% file: appendices/appendix_corl.tex
\section{Platform and Task Setup}
\label{app:setup}
\textbf{Hardware.} The platform is a Franka Research~3 arm with a
five-fingered Tesollo DG5F hand. A RealSense D455 RGB-D camera is mounted on
the wrist; its colour stream ($848{\times}480$) is the RGB observation, and
point clouds of 2048 points are reconstructed from its depth stream. Two-view
variants add the image of a second, static third-person camera that looks at
the work area. The robot-facing action is 10-dimensional: end-effector translation (3),
orientation (3, Euler angles) and four hand-joint commands. When a 6D
rotation~\cite{zhou2019continuity} is used inside the policy, state and action
grow to 13 dimensions while the robot interface stays 10-dimensional.

\textbf{Demonstrations.} An operator teleoperates the arm with an HTC Vive
tracker and the hand with a Manus glove. A ROS\,2 recorder time-synchronises
hand, robot and camera streams at 30\,Hz. Each demonstration is segmented into
motion primitives and the two learned skills, and each skill episode is stored
separately. Policies run at 10\,Hz: replaying the same demonstrations at
30\,Hz and at 10\,Hz gave exactly the same successes and failures (4/11).

\textbf{Task and start positions.} The RJ45 cable connector sits in a socket
on a jig. Demonstrations were recorded with the jig in five placements
(centre, front, back, left and right). All reported rollouts use two of them:
the \emph{centre} jig, with the robot starting at a lateral offset $y{=}0$\,m,
and the \emph{left} jig, with $y{=}{+}0.05$\,m (occasionally
${+}0.04$--$0.045$\,m). At run time FoundationPose~\cite{wen2024foundationpose}
identifies which of the two jigs holds the cable, classical motion moves the
robot to a preset start pose for that jig, and the learned skills take over. The \emph{grasp} skill closes the hand on
the connector and pulls it out of its socket; it lifts only once the hand's
grip effort exceeds a threshold (mostly 80, tuned between 10 and 120 over the
project). Classical motion then carries the connector to a third jig with the
target socket, where the \emph{insert} skill seats it.

\textbf{Outcome stages.} Each rollout is scored by the furthest of four nested
stages it reached. \emph{Fail} ($F$): the hand did not close on the connector.
\emph{Grasp} ($G$): the hand closed on the connector but never reached the grip
threshold, so the connector stayed in its socket; unlike a fail, the grasp
skill found the connector. \emph{Pick} ($P$): the connector was pulled out but
not seated. \emph{Insert} ($I$): the connector was seated in the target
socket. Because the stages are nested, an inserted trial was necessarily
picked first, which is why the pick rate counts both inserted and picked
trials, $(I{+}P)/n$. The insert-given-pick rate $I/(I{+}P)$ judges an insert
skill only on the trials in which it received a connector.

\section{Datasets and Terminology}
\label{app:data}
\begin{figure}[h]
\centering
\includegraphics[width=0.62\linewidth]{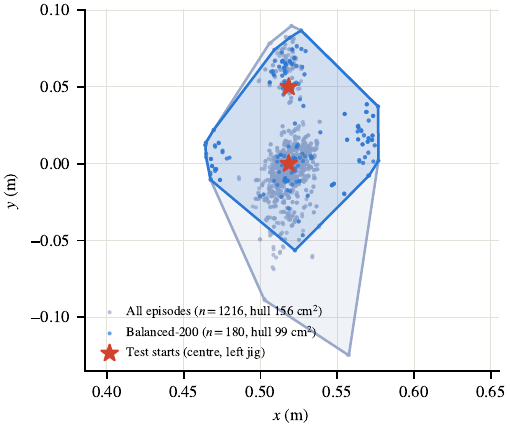}
\caption{Start positions ($x$, $y$ of the end effector in the robot base
frame) of the grasp episodes in the all-episode and balanced-200
sets, with their convex hulls. Stars mark the two test start positions. The
balanced set keeps the spread of start positions while evening out the
clusters formed by the jig placements.}
\label{fig:app-hull}
\end{figure}

Table~\ref{tab:app-data} lists the datasets referred to in the paper, and
Fig.~\ref{fig:app-hull} shows the spread of their start positions. \emph{Full} (or \emph{big}) sets hold
all demonstrations of a skill and are unevenly spread over the five jig
placements. \emph{Balanced} sets are subsets with equal numbers of episodes per
start-pose cluster, so that the less-demonstrated left jig is not
under-represented. The grasp skill is trained on the balanced set and the
insert skill on all episodes: grasp success depends strongly on the approach
pose, so evening out the start-pose clusters helps most there, whereas every
insertion targets the same socket, which makes start-pose balancing irrelevant
and lets the harder, millimetre-precision seating benefit instead from the full
data volume. \emph{Recovery episodes} are demonstrations that correct a
misaligned connector before completing the skill, so that the policy sees how
to recover. \emph{Light-variation episodes} were recorded under changed lighting and
backgrounds (see the campaign below).

\begin{table}[h]
\centering
\caption{Datasets, with grasp / insert episode counts.}
\label{tab:app-data}
\small
\setlength{\tabcolsep}{3pt}
\begin{tabular}{@{}p{0.44\linewidth}p{0.2\linewidth}p{0.3\linewidth}@{}}
\toprule
Dataset & Episodes & Used for \\
\midrule
\multicolumn{3}{@{}l}{\emph{Image-based skills}} \\
Full grasp set & 1472 & RGB grasp skills \\
Balanced-186 & 186 & RGB DP, $\pi_{0.5}$ grasp \\
Small insert set $+$ recovery & 294 & RGB DP, $\pi_{0.5}$ insert \\
Two-view balanced-200 $+$ recovery & 180 $+$ 108 (insert) & two-view inserts \\
\midrule
\multicolumn{3}{@{}l}{\emph{Point-cloud skills}} \\
All episodes & 1216 / 1185 & \\
\ $+$ recovery & 1322 / 1291 & iDP3 insert skills \\
\ $+$ light variation & 1610 / 1577 & \\
\ $+$ recovery $+$ light variation & 1713 / 1683 & flow-matching PCD skills \\
Balanced-200 & 180 / 166 & iDP3 grasp skills \\
\ $+$ recovery & 286 / 272 & iDP3 grasp skills \\
\ $+$ light variation & 574 / 558 & \\
\ $+$ recovery $+$ light variation & 680 / 664 & PointNet DP3 skills \\
Recovery episodes only & 106 / 106 & \\
Light-variation episodes only & 394 / 392 & \\
\bottomrule
\end{tabular}
\end{table}

\textbf{Recovery and light-variation campaign.} The recovery and
light-variation episodes were collected in one campaign:
centre-jig grasps (25 episodes) and left-jig grasps (25); recovery episodes
with an improper grasp that is corrected before insertion (50) and with a
correct grasp followed by a corrected, initially improper insertion (50);
centre- and left-jig grasps with the backlight off (25 each); and a set under
changing conditions with five episodes per condition, plus one extra episode
per condition kept as a test set. That set combines
three factors: the LED lamp off, or on at different places spread along a
circle around the table, together with the overhead light off or on; the
background as usual, with the screen removed so that people and objects are
visible, or with a screen at the end of the table covering the pole of the
static third-person camera; and either only the centre jig on the table, or
both jigs with the cable in the centre jig.

\textbf{Point-cloud versions.} Every point-cloud set exists in two
\emph{reference frames}: the \emph{camera frame}, in which points are expressed
relative to the moving wrist camera, and the \emph{robot-base frame}, in which
the same points are transformed into the fixed base of the robot using the
hand--eye calibration. The
\emph{region-of-interest (ROI) crop} keeps only the points inside a 3D box
around the region where the hand acts, the connector for the grasp skill and
the target socket for the insert skill, and then samples 2048 points with
farthest-point sampling. ``V2'' and ``refined'' clouds are re-processed
exports of the same recordings.

\section{Evaluation Protocol and Statistics}
\label{app:protocol}
\textbf{Sessions and trial counts.} A test session runs one grasp-then-insert
pairing five times from each of the two start conditions ($n{=}10$). Promising
checkpoints were re-tested in later sessions, with other partner skills and
under the robustness conditions, whereas clearly failing configurations were
often stopped after two or three trials. The standardised rates pool every
trial of a checkpoint across sessions and pairings, so $n$ varies: for example
the $\pi_{0.5}$ grasp skill was run 117 times, the best iDP3 grasp skill 73
times and the robot-frame PointNet grasp of the frame comparison 14 times,
while its camera-frame counterpart was run in two sessions (20 trials). Rates
are therefore always given with their trial counts or 95\% Clopper--Pearson
intervals.

\textbf{Matched comparisons.} For every before/after pair we compared the two
runs' saved training configurations key by key and label the pair with every
setting that differed. Pairs that differ only in the named setting are called
\emph{isolated}; the others are shown as \emph{several changes}, with each
difference listed. Before/after success is compared with a two-sided Fisher
exact test.

\textbf{Robustness conditions.} \emph{Light variation} changes the lighting
(lights off, LED lamp on or moved, backlight off); \emph{background variation}
removes or moves the screen behind the work area;
\emph{new environment} moves the robot to a different site; \emph{sunlight}
exposes the work area to direct sunlight; the \emph{tight-tolerance socket}
replaces the target socket by one with a smaller clearance.
\ifdefined\robustinmain
Table~\ref{tab:robust} in the main text gives the results under each condition.
\else
Table~\ref{tab:app-robust} gives the results under each condition. RGB DP
pipelines degrade where the point-cloud pipeline does not: under changed
lighting they kept picking but inserted far less often, and at the new site
the RGB pipeline rarely picked.

\begin{table}[h]
\centering
\caption{Robustness. Top: the iDP3 point-cloud pipeline, insert successes out
of five trials at the centre and left jig; rows marked $^{\ddagger}$ use the
large-UNet grasp and batch-320 insert checkpoints of the same recipe. Bottom:
RGB DP pipelines, picks, inserts / trials pooled over both jigs.}
\label{tab:app-robust}
\small
\setlength{\tabcolsep}{5pt}
\begin{tabular}{@{}p{0.5\linewidth}cc@{}}
\toprule
Condition & Centre & Left \\
\midrule
Nominal                               & 5/5 & 5/5 \\
Light variation                       & 5/5 & 5/5 \\
Background variation                  & 5/5 & 4/5 \\
New environment                       & 5/5 & 4/5 \\
New environment $+$ light variation   & 3/5$^{\dagger}$ & 4/5 \\
Sunlight$^{\ddagger}$                 & 4/5 & 4/5 \\
Sunlight $+$ light variation$^{\ddagger}$ & 4/5 & 5/5 \\
Tight-tolerance socket$^{\ddagger}$   & 2/5 & 0/5 \\
\midrule
& Nominal & Perturbed \\
RGB DP, from scratch; lighting        & 6, 4 / 10 & 10, 1 / 16 \\
RGB DP, ImageNet; lighting            & 10, 9 / 10 & 13, 8 / 13 \\
RGB DP, 6D; new site, left jig        & -- & 1, 0 / 3 \\
\bottomrule
\multicolumn{3}{@{}l}{\footnotesize $^{\dagger}$4/5 after re-tuning the insert start position for the new site.}
\end{tabular}
\end{table}
\fi

\section{Policies, Codebases and Settings}
\label{app:policies}
\textbf{Codebases.} DP, DP3, iDP3, flow matching and ACT are trained in one
PyTorch Lightning codebase for multi-GPU training, whose policy package is
adapted from the 3D Diffusion Policy code of Ze et~al.~\cite{ze2024dp3}. The
multi-stage point encoder is adapted from the iDP3 code~\cite{ze2024idp3}, the
flow-matching policy reuses the same encoders and 1D UNet with a conditional
optimal-transport flow-matching objective~\cite{lipman2023flow}, and ACT is
vendored from the original implementation~\cite{zhao2023act}. Our additions
are the Lightning training loop with distributed data parallelism and MLflow
logging, 6D rotation and relative (temporally normalised) action options, the
point-cloud frame conversion, ROI cropping and sampling pipeline, a two-view
RGB encoder, the spotlight augmentation, and a corrected last-checkpoint
callback. $\pi_{0.5}$~\cite{pi05} is finetuned with a fork of the openpi code
(JAX) that adds a data loader for our recordings and options for 6D rotation,
relative actions, temporal normalisation and a real second camera image. The
equivariant policy uses the EquiBot code~\cite{yang2024equibot} in its own
container.

\begin{table}[h]
\centering
\caption{Default settings of each family's best configuration ($H$:
prediction horizon; $n_\text{a}$: executed steps). Unless noted, grasp skills
are trained on the balanced set at batch 32 per GPU and insert skills on all
episodes at batch 160, on four H100s.}
\label{tab:app-policies}
\small
\setlength{\tabcolsep}{2.5pt}
\begin{tabular}{@{}p{0.2\linewidth}p{0.44\linewidth}p{0.32\linewidth}@{}}
\toprule
Family & Encoder and head & Training \\
\midrule
DP (RGB) & ResNet with one block per stage, spatial softmax; 1D UNet, DDIM; $H{=}4$, $n_\text{a}{=}4$ & 1000 epochs; insert skill also at batch 32 (small insert set) \\
DP3 & PointNet (LayerNorm in its layers), output dim 64; UNet, DDIM; $H{=}4$, $n_\text{a}{=}3$ & 1000 epochs; both skills at batch 160 on the balanced set \\
iDP3 & multi-stage PointNet, output dim 128; UNet, DDIM; $H{=}16$, $n_\text{a}{=}8$ & 1000--3000 epochs \\
Flow matching & ResNet, PointNet or multi-stage encoder; UNet; 5--10 integration steps & 100--1000 epochs; both skills at batch 160 \\
ACT & ResNet backbone, CVAE with a transformer decoder & 150 epochs; both skills at batch 128 \\
$\pi_{0.5}$ & PaliGemma 2B with a 300M action expert, LoRA; action horizon 50, chunks of 10 executed & 30k steps per skill, batch 32 \\
\bottomrule
\end{tabular}
\end{table}

\section{Changes Tested in the Ablation Study}
\label{app:changes}
Table~\ref{tab:app-changes} lists every setting we varied. A before/after
comparison takes two policies that share all other training settings and
differ in one of these rows; where a comparison changes several rows at once,
the figures and tables of the paper name all of them.

\begin{table}[h]
\centering
\caption{Settings varied in the ablation study, with the values tried.}
\label{tab:app-changes}
\small
\setlength{\tabcolsep}{3pt}
\begin{tabular}{@{}p{0.3\linewidth}p{0.66\linewidth}@{}}
\toprule
Setting & Values tried \\
\midrule
\multicolumn{2}{@{}l}{\emph{Data}} \\
Episode selection & all demonstrations; subset balanced across start positions \\
Extra episodes & recovery episodes; light-variation episodes \\
Cameras & wrist only; wrist and static third-person camera \\
\multicolumn{2}{@{}l}{\emph{Observation processing}} \\
Image augmentation & none; spotlight; spotlight and noise injection; plus colour jitter; plus cutout \\
Point-cloud augmentation & SE(3) jitter, translation or rotation of the cloud; perturbed start pose \\
Image-encoder initialisation & from scratch; ImageNet; R3M; frozen or finetuned at a lower learning rate \\
Image encoder & ResNet with one block per stage; ResNet-34; TinyViT (offline only); spatial-softmax or mean pooling \\
Point-cloud frame & camera; robot base \\
Point-cloud normalisation & per-cloud normalisation on or off; LayerNorm after the output or after every layer of the PointNet \\
Point-cloud crop & none; region-of-interest box around the connector; box around the socket; gripper-centred or fixed box; 1000 or 2048 points \\
\multicolumn{2}{@{}l}{\emph{Action representation}} \\
Action type & absolute; relative (temporally normalised); delta \\
Rotation & Euler angles; 6D~\cite{zhou2019continuity} \\
Horizon $\times$ executed steps & $4{\times}3$, $8{\times}4$, $16{\times}8$, $32{\times}16$; $\pi_{0.5}$ chunks of 3 or 10 \\
\multicolumn{2}{@{}l}{\emph{Network and training}} \\
Network size & UNet widths $[256,512,1024]$ or $[512,1024,2048]$; step embedding 96, 128 or 256; encoder output 64, 128 or 256; hidden size 256; 4, 5 or 6 point-encoder layers \\
Optimiser & learning rate $5{\times}10^{-5}$ to $3{\times}10^{-4}$; weight decay $10^{-6}$ or $10^{-4}$; 500, 1500 or 2000 warm-up steps; dropout 0 or 0.1; batch 32, 128, 160 or 320; 300, 1000 or 3000 epochs \\
Sampler & DDIM or DPM-Solver++; sample, $\epsilon$ or $v$ prediction; 5 or 10 flow-matching integration steps \\
VLA finetuning & LoRA; full finetuning \\
\multicolumn{2}{@{}l}{\emph{Task structure and family}} \\
Skill split & separate grasp and insert policies; one grasp-and-insert policy \\
Policy family & DP, DP3, iDP3, flow matching (three encoders), ACT, $\pi_{0.5}$, equivariant \\
\bottomrule
\end{tabular}
\end{table}

\section{Offline-Metric Analysis}
\label{app:offline}
\begin{figure}[t]
\centering
\ifdim\columnwidth>3.6in\includegraphics[width=5.2in]{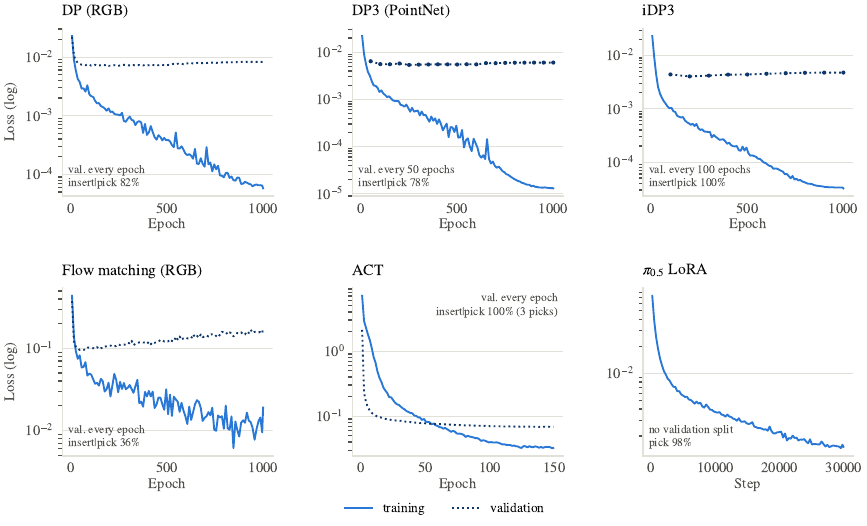}\else\includegraphics[width=\columnwidth]{figures/fig_loss_curves_family_serif.pdf}\fi
\caption{Training loss (solid) and validation loss (dotted) of the deployed
skill of each policy family, from the runs' training logs; each panel states
how often the validation loss was computed and the skill's robot rate. In every
family the validation loss reaches its minimum early and is then flat or rising
while the training loss keeps falling, regardless of the robot rate
(36--100\% insert-given-pick); $\pi_{0.5}$ finetuning uses no validation
split.}
\label{fig:app-loss}
\end{figure}
\textbf{Metrics.} For each checkpoint the policy is replayed on the validation
episodes held out from its own training set: at 50 evenly spaced frames of
each episode it receives the recorded observation and predicts an action
chunk. From these predictions we compute 67 metrics in five groups:
validation loss; open-loop errors against the demonstrated actions
(first-step and horizon error, Fr\'echet and displacement distances, MMD and
KL between predicted and demonstrated action distributions); full-episode
metrics of the predicted end-effector path (goal-region error, reach rates
within 1, 2 and 5\,mm, path length and path-length ratio, speed, jerk,
spectral power, smoothness, drift); encoder statistics (keypoint jitter, feature
cosine similarity, saliency); and denoising confidence.

\textbf{The 67 metrics and how they were chosen.} We used every metric that our
offline evaluation tool computes on a checkpoint's validation split, and left
out only bookkeeping columns (epoch, step count, parameter count); no metric
was chosen or dropped by looking at robot results. Most are standard: the
displacement, Fr\'echet and distribution distances (MMD~\cite{gretton2012mmd},
KL) come from trajectory-prediction and imitation-learning evaluation, and
jerk, path-length ratio and the spectral arc length
(SPARC~\cite{balasubramanian2015sparc}) are standard measures of movement
smoothness, and the RoboEval benchmark~\cite{wang2025roboeval} scores
simulated manipulation policies with the same kinds of behaviour metrics (path
length, smoothness, stage-wise progress) alongside success counts; the encoder
and denoising statistics are diagnostics of our own.
Table~\ref{tab:app-metrics} lists them. The first 38 come from replaying
predictions along each validation episode; the other 29 from a second tool
that reports the same kinds of error per axis and a composite score.

\begin{table}[h]
\centering
\caption{The 67 offline metrics, by group (count in brackets).}
\label{tab:app-metrics}
\small
\setlength{\tabcolsep}{3pt}
\begin{tabular}{@{}p{0.3\linewidth}p{0.66\linewidth}@{}}
\toprule
Group & Metrics \\
\midrule
Training (1) & best validation loss \\
Open-loop error (10) & first-step position error, horizon error growth, average displacement error, cumulative position error, Fr\'echet distance (position; all dimensions), MMD and KL between predicted and demonstrated actions, consistency of overlapping action chunks, sensitivity to small state perturbations \\
Uncertainty (4) & spread of actions and of positions over 8 denoising samples, denoising confidence, denoising loss \\
Encoder (4) & keypoint jitter, number of peaked keypoints, cosine similarity of consecutive features, input saliency \\
Goal region (10) & mean and median position error, median goal error, mean and median orientation error, reach within 1, 2 and 5\,mm, translational and rotational drift \\
Motion quality (9) & jerk, path-length ratio, dimensionless-jerk smoothness, SPARC, mean speed, Cartesian and orientation path length, spectral power of the predicted path and its ratio to the demonstration \\
Second tool (29) & path and goal position error (total and per axis, median, spread), path and goal orientation error (total and per axis), reach within 2, 5 and 10\,mm (position; position and orientation), jerk, path-length ratio, translational and rotational drift, a composite score \\
\bottomrule
\end{tabular}
\end{table}

\textbf{Scopes and correlation.} A \emph{scope} is the set of checkpoints
whose metrics are correlated together with their robot success rates. We use
four scopes of diffusion-family checkpoints (RGB DP with PointNet DP3, iDP3
alone, all three, and RGB DP with iDP3), and split each by jig coverage:
checkpoints tested only at the centre jig (80, mostly short hyper-parameter
probes) and checkpoints tested at both jigs (120). Within a scope each metric
is correlated (Pearson $r$, episode-weighted) with the pick rate of grasp
checkpoints or the insert-given-pick rate of insert checkpoints. A cell counts
as reliable only if it survives Benjamini--Hochberg correction at
$q{=}0.05$ over its matrix, keeps its sign when any single policy is left
out, and has at least ten checkpoints.

\textbf{Loss curves.} Fig.~\ref{fig:app-loss} shows why validation loss
does not rank checkpoints, with the deployed skill of each family. The
validation loss reaches its minimum within the first 250 epochs in every
family and then rises by 12--26\% for DP, PointNet DP3 and iDP3, more than
doubles for flow matching (so for all eight flow-matching skills with robot
trials) and stays flat for ACT, while the training loss keeps falling by two to
three and a half orders of magnitude. The curves of skills that inserted in
78--100\% of the trials in which they received the connector (DP, DP3, iDP3,
ACT) and of one that inserted in 36\% (flow matching) differ in their loss
scale, not in their shape, and every deployed checkpoint is the final epoch,
far past the validation minimum. Over the 21 iDP3 insert checkpoints behind
Fig.~\ref{fig:results}b the best validation loss is uncorrelated with the
insert-given-pick rate ($r{=}{+}0.08$, $p{=}0.72$), the grey ``Validation
loss'' cell of the heatmap in the main text.